\documentclass[letterpaper, 10 pt, conference]{ieeeconf}  % Comment this line out if you need a4paper

\IEEEoverridecommandlockouts                              % This command is only needed if 
\usepackage{graphics} % for pdf, bitmapped graphics files
\usepackage{epsfig} % for postscript graphics files
\usepackage{mathptmx} % assumes new font selection scheme installed
\usepackage{times} % assumes new font selection scheme installed
\usepackage{amsmath} % assumes amsmath package installed
\usepackage{mathptmx}
\usepackage{mathrsfs}
\usepackage{todonotes}
\usepackage{balance}
\DeclareMathAlphabet{\mathcal}{OMS}{cmsy}{m}{n}
\usepackage{xcolor}
\usepackage{caption}
\usepackage{algorithm}
\usepackage{algorithmic}
\usepackage[algo2e]{algorithm2e} 
\usepackage{hyperref}
\title{\LARGE \bf
An Efficient Model-Based Approach on Learning Agile Motor Skills without Reinforcement
}

\usepackage{color}

\newcommand{\shj}[1]{{ \color{black}#1}}

\author{Haojie Shi$^{1,2*}$, Tingguang Li$^{1*}$, Qingxu Zhu$^{1}$, Jiapeng Sheng$^{1}$, Lei Han$^{1}$ and Max Q.-H. Meng$^{3\dag}$, \textit{Fellow, IEEE} 
\thanks{$\dag$ Corresponding author.}%
\thanks{$*$ Equal Contribution.}%
\thanks{$^{1}$ Haojie Shi, Tingguang Li, Qingxu Zhu, Jiapeng Sheng, and Lei Han are affiliated with Tencent Robotics X, China, (email: {\tt\footnotesize{haojieshi, teaganli,qingxuzhu,kevinsheng,lxhan}@tencent.com})}%
\thanks{$^{2}$ Haojie Shi is from the Chinese University of Hong Kong, and this work was done during internship at Tencent Robotics X Lab.}%
\thanks{$^3$ Max Q.-H. Meng is with Shenzhen Key Laboratory of Robotics Perception and Intelligence and the Department of Electronic and Electrical Engineering at Southern University of Science and Technology in Shenzhen, China. He is a Professor Emeritus in the Department of Electronic Engineering at The Chinese University of Hong Kong in Hong Kong and was a Professor in the Department of Electrical and Computer Engineering at the University of Alberta in Canada. (email: {\tt\footnotesize max.meng@ieee.org})}%
}
\begin{document}

\maketitle
\thispagestyle{empty}
\pagestyle{empty}

%%%%%%%%%%%%%%%%%%%%%%%%%%%%%%%%%%%%%%%%%%%%%%%%%%%%%%%%%%%%%%%%%%%%%%%%%%%%%%%%
\begin{abstract}
Learning-based methods have improved locomotion skills of quadruped robots through deep reinforcement learning. However, the sim-to-real gap and low sample efficiency still limit the skill transfer. To address this issue, we propose an efficient model-based learning framework that combines a world model with a policy network. We train a differentiable world model to predict future states and use it to directly supervise a Variational Autoencoder (VAE)-based policy network to imitate real animal behaviors. This significantly reduces the need for real interaction data and allows for rapid policy updates. We also develop a high-level network to track diverse commands and trajectories. Our simulated results show a tenfold sample efficiency increase compared to reinforcement learning methods such as PPO. In real-world testing, our policy achieves proficient command-following performance with only a two-minute data collection period and generalizes well to new speeds and paths. The results are shown in \href{https://youtu.be/2R4RffrzS98}{$youtu.be/2R4RffrzS98$}.
% We initially train the policy within a simulation environment and subsequently fine-tune it using a physical robot. Simulated results show a tenfold sample efficiency increase compared to a model-free reinforcement learning baseline. Transitioning to real-world testing, our policy achieves proficient command-following performance with only a two-minute data, and generalizes well to new speeds and paths.

\end{abstract}

%%%%%%%%%%%%%%%%%%%%%%%%%%%%%%%%%%%%%%%%%%%%%%%%%%%%%%%%%%%%%%%%%%%%%%%%%%%%%%%%
\section{INTRODUCTION}
Learning-based methods~\cite{da2020learning,lee2020learning,rudin2022learning,yang2020multi,yang2022fast,shi2023terrain} have recently demonstrated significant advantages in acquiring agile motor skills for quadrupedal robots. In particular, model-free deep Reinforcement Learning (RL) algorithms enables them to mimic animal motions \shj{so as to attain natural and agile motor skills}~\cite{peng2018deepmimic,peng2020learning,li2023learning,escontrela2022adversarial,zhu2023ncp,han2023lifelike}.

However, model-free RL algorithms~\cite{schulman2015trust,schulman2017proximal,haarnoja2018soft} usually require substantial on-policy data to improve their performance. Given the cost of collecting data in simulation compared to the real world, these algorithms often train policies in simulation and then deploy them on physical robots through zero-shot transfer. However, the policies learned in simulation may not consistently perform well in real-world scenarios due to the persistent sim-to-real gap. Researchers have attempted to mitigate this gap using techniques like domain randomization \cite{tobin2017domain} and domain adaptation within simulation environments to enhance policy robustness. Nevertheless, these techniques do not provide a fundamental solution and cannot guarantee successful transfer. \cite{xie2021dynamics} argues that dynamics randomization and adaptation approaches may not consistently address sim-to-real transfer challenges, leaving the sim2real gap unresolved.
% \footnotetext[1]{We provide a video to show the results in  \href{https://youtu.be/mjIxrDC_QtQ}{$youtu.be/mjIxrDC_QtQ$}.}
\begin{figure}[t]
	% \vspace{-.3cm}
	\centering
	\includegraphics[width=0.45\textwidth]{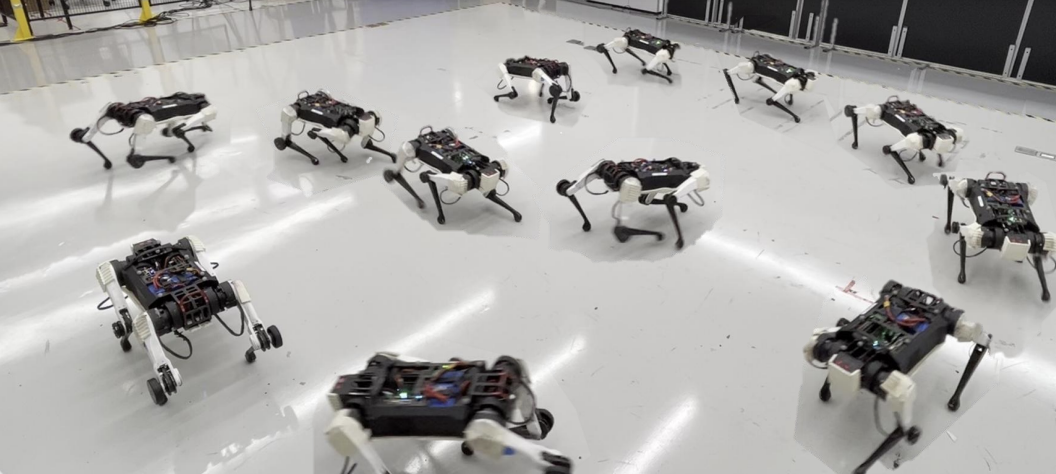}
	\caption{Our robot Max follows the U-shape path after fine-tuned in the real world.}
	\label{fig:real_show}
	\vspace{-.4cm}
\end{figure}
On the other hand, an alternative approach is to train or fine-tune the policy directly on a real robot, which can effectively address the problem. \cite{smith2022legged} utilizes a model-free off-policy reinforcement learning algorithm for policy fine-tuning in the real world, albeit it still necessitates more than 2 hours data for fine-tuning. To increase sample efficiency, \cite{wu2023daydreamer} adopts a model-based reinforcement learning approach, which enables direct policy training on the real robot. Nevertheless, since the policy network is trained through a model-free reinforcement learning algorithm, this method still requires over one hour to train a basic policy for walking towards predefined directions.

In the realm of computer graphics, ControlVAE \cite{yao2022controlvae} has demonstrated superior sample efficiency compared to deep reinforcement learning. It achieves this by co-training a world model with a VAE-based policy network~\cite{kingma2013auto}. Building on this concept, we introduce a model-based learning framework to close the sim2real gap by directly fine-tuning policies on real robots. First we train a world model capable of predicting several consecutive states of the robot. Leveraging the differentiability of the world model, we can train an end-to-end control policy by direct backpropagation. This policy imitates reference trajectories obtained from real dogs by interacting with the world model. Additionally, we develop a high-level policy for generating latent variable within the VAE \cite{kingma2013auto}. This empowers the robot to follow various high-level commands and track diverse paths.

In simulated experiments, our method exhibits a tenfold improvement in sample efficiency compared to PPO~\cite{schulman2017proximal}, both during training and adaptation. In real robot experiments, our policy effectively tracks a oblong path at speeds of 0.6m/s, 0.9m/s, and 1.2m/s with just 2 minutes of fine-tuning. Furthermore, we also evaluate our policy generalization ability in new speed commands and unseen paths, highlighting our method's robust generalization capability.

In conclusion, the main contributions of this paper are:
\begin{enumerate}
	\item We present a model-based learning framework to acquire agile skills in quadrupedal robots within simulations and fine-tune them on real robots, substantially enhancing the sample efficiency of learning-based methods in the robotics domain.
	\item We assess our approach in both simulation and the real robot, demonstrating that with only 2 minutes of fine-tuning, our robot effectively executes reference commands.
	\item We establish the generalization capability of our method with real robot experiments, as the fine-tuned policy follows previously unseen commands and paths.
\end{enumerate}

\section{RELATED WORK}

As model-free deep reinforcement learning algorithms \cite{schulman2015trust,schulman2017proximal,haarnoja2018soft} continue to advance rapidly, recent research has achieved notable success in training expert policies for quadrupedal locomotion. In contrast to classical control methods \cite{bellicoso2017dynamic,bledt2018cheetah,carius2018trajectory,carius2019trajectory,di2018dynamic,fankhauser2018robust,carpentier2018multicontact,aceituno2017simultaneous,winkler2018gait,farshidian2017efficient}, learning-based approaches harness the power of deep neural networks to acquire agile motor skills. Notably, \cite{peng2018deepmimic,peng2020learning} employ PPO~\cite{schulman2017proximal} to emulate natural motion patterns observed in animals. Moreover, \cite{li2023learning} introduces a novel approach that incorporates terrain information while mimicking the behavior of real animals.

However, model-free deep reinforcement learning algorithms demand an enormous volume of interaction data, making it infeasible to collect on a real robot. Consequently, they often train the policy in simulation and attempt zero-shot transfer into the real world. To address the sim2real gap, \cite{peng2020learning} introduces an environmental encoding approach, optimizing it on the real robot by maximizing total returns for swift adaptation. It's crucial to note that the effectiveness of this adaptation strategy hinges on the degree of similarity between the simulation and real-world environments and may not be universally applicable across all tasks. Furthermore, \cite{li2023learning,tan2018sim,shi2022reinforcement,tobin2017domain} employ domain randomization techniques to cultivate a robust policy and employ domain adaptation to address the sim2real gap. \shj{\cite{radosavovic2023real} reduces the sim2real gap by utilizing pre-trained representations that prove effective across various real-world robotic tasks.} While these methods can alleviate the impact of the sim2real gap, they do not offer a fundamental solution. Notably, \cite{xie2021dynamics} demonstrates that their policy can be transferred to the real robot without necessitating domain randomization, questioning the necessity of this technique. In summary, the challenge of transferring model-free reinforcement learning policies from simulation to reality remains an open problem.

To increase sample efficiency for reinforcement learning, recent years have witnessed great progress in model-based reinforcement learning algorithms \cite{luo2018algorithmic,clavera2018model,kurutach2018model}. They first learn a dynamics model in the simulation and then improve their policy using model-free RL with imagined data produced by the learned model. To increase the prediction power of the world model, further research focuses on learning a compact latent space of world model \cite{hafner2020mastering,hafner2019dream,hafner2023mastering}, and also succeeds in the real robot training \cite{wu2023daydreamer}. In this way, the sim2real gap does not exist since they train the policy directly in the real robot. While they can train walk policy in a real robot in one hour, it is still not evaluated how much data it will take to train a more complex policy like imitating an animal or following a desired path in our task. And since the policy is trained by the model-free reinforcement learning algorithm, the sample efficiency is still limited. Meanwhile, training directly in the real robot from scratch fails to take advantage of the simulation environment. In contrast, our method trains both the world model and control policy in a supervised manner, resulting in significantly enhanced sample efficiency. Additionally, we adopt a two-stage approach that involves training the policy in simulation to create a warm-up policy, followed by fine-tuning it in the real world using just two minutes of data. This significantly reduces the amount of real-world data required and enables the learning of more sophisticated motor skills.

ControlVAE~\cite{yao2022controlvae} is an innovative technique in computer graphics that utilizes a VAE-based policy, supervised by a differentiable world model. This approach provides significantly higher sample efficiency than deep reinforcement learning algorithms, but it mainly concentrates on policy training for human motion generation within a simulation context. To expand on this concept, our proposed framework combines world model and policy learning in a supervised manner, resulting in a learning framework that enhances training efficiency during the fine-tuning stages on a real robot with a regularization term. Consequently, our approach allows for deployment on a real quadrupedal robot system with only a 2-minute fine-tuning period.

\section{METHODOLOGY}
\begin{figure}[t]
	\centering
	\includegraphics[width=0.45\textwidth]{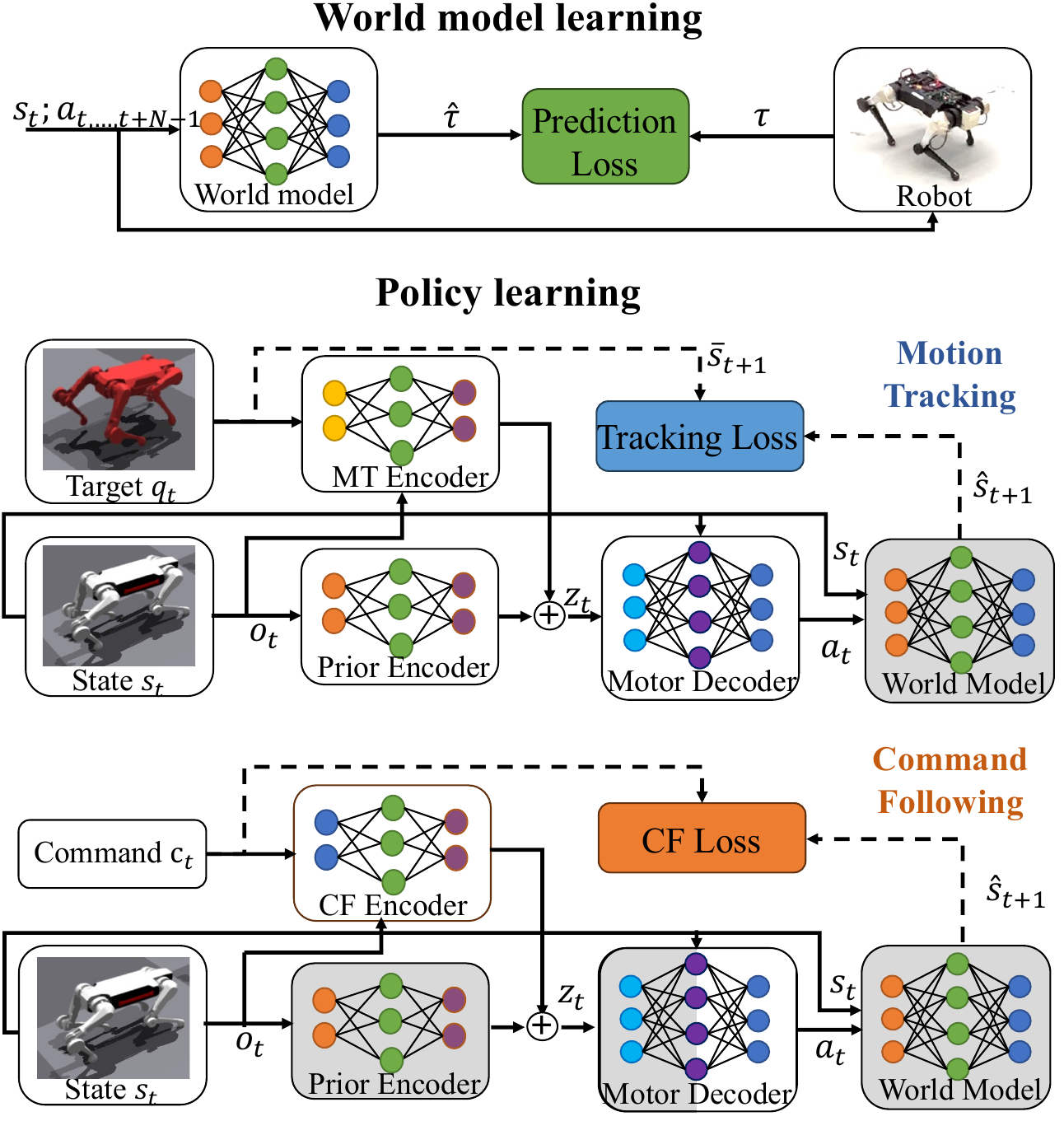}
	\caption{Overview of our learning framework. The gray block represents fixed parameters. For the command following task, the Motor Decoder is fixed when training from scratch and becomes trainable during real-world fine-tuning.}
	\label{fig:framework}
	\vspace{-.4cm}
\end{figure}

Our framework contains two parts, i.e. a world model and a control policy, as shown in Fig.~\ref{fig:framework}. 
The world model learns to approximate the unknown dynamics of the simulation and the reality. Given current robot state and action, it predicts next state.
The control policy learns agile behaviors by tracking motions from real animals. Instead of interacting with a simulator, it directly collects samples predicted by the trained world model. 
% As for policy learning, we first train the imitation task with a VAE-based \cite{kingma2013auto} network that learns to imitate the behavior of real animals, and then train a command network to find an appropriate latent code given the current command for the tracking task.
Both the world model and the control policy are updated in a supervised  manner and trained iteratively: we first collect state-action pairs under a fixed control policy to fit the system dynamics using the world model. Then the control policy is updated by interacting with the fixed world model. The whole process repeats until the control policy converges. 
% During the world model learning phase, the policy network is fixed, and verse vice.

\subsection{World Model Learning}
We commence by training the world model $f_{\text{w}}$. It predicts the next state based on the current state and action, utilizing a residual form as follows:
\begin{equation}
    \hat{s}_{t+1} =f_{\text{w}}(\Delta s_t | o_t,a_t)+s_t,
    \label{eq:prediction}
\end{equation}
where $f_{\text{w}}(\Delta s_t | o_t,a_t)$ is a neural network parameterized by $\theta_\text{w}$, $s_t$ represents the robot state at time $t$, encompassing robot position, orientation, linear velocity, angular velocity, joint positions and joint velocities. $o_t$ corresponds to the robot observation, including robot linear velocity, angular velocity, joint position, and joint velocity at robot local frame. $a_t$ is the target angle for each joint which can be converted to joint torques through a PD controller. $\hat{s}_{t+1}$ represents the predicted state at the next time step $t+1$. 
When training the world model, the robot interacts with the simulator under a fixed control policy to collect state-action sequences $\tau=\{s_0,a_0,s_1,a_1, ...,s_n,a_n\}$. The world model is trained in supervised learning manner with the n-step prediction loss that is conducive to prediction in a long time horizon:
\begin{gather}
    L^{\text{w}}_t = \sum_{t=1}^{n}{\lVert \hat{s}_{t}-s_t \rVert},
    % s.t.\quad \hat{s}_{t+1} = \theta_{wm}(o_t,a_t)+\hat{s}_t, s_{t+1} = P(s_t,a_t)
    \label{eq:prediction_loss}
\end{gather}
where $\hat{s}_t$ is the predicted robot state and $s_t$ represents the ground truth robot state either from simulator or from the real robot during the fine-tuning stage. 
% denotes the dynamics system, with $a_t$ being generated by the current policy, and the initial state  is set to equal $s_0$. Subsequently, we can achieve a precise approximation of the dynamics system for the upcoming frames. 

\subsection{Motion Tracking}
In the context of the motion tracking task, our objective is to imitate motion sequences collected from real animals. We formulate this problem into an encoder-decoder architecture, where we encode the reference motion sequence into a latent embedding and decode this latent embedding together with robot observation into joint motor action.
We take a VAE-based~\cite{kingma2013auto} architecture with an Motion Tracking Encoder $\pi_{\text{MT}}(z_t|o_t, \textbf{\textit{q}}_t)$ to encode the observation $o_t$ and a sequence of future reference motions $\textbf{\textit{q}}_t$ into a latent variable $z_t$. The Motor Decoder $\pi_{\text{M}}(a_t|o_t, z_t)$ takes $o_t$ and $z_t$ and produces the action $a_t$. 
% With the world model, the next state $\hat{s}_{t+1}$ can be reconstructed. 
Besides, we incorporate an additional state-conditional prior $\pi_{\text{prior}}(z_t|o_t)$ to disentangle distinct skills within the latent space, as emphasized in~\cite{yao2022controlvae}.
% As emphasized in \cite{yao2022controlvae}, the state-conditional prior is advantageous for disentangling distinct skills within the latent space. Consequently, we have incorporated this technique into our approach. 
% Given the robot observation $o_t$, the prior distribution of latent code $z_t$ can be sampled from the state-dependent gaussian distribution:
% \begin{equation}
%     p(z_t|o_t) \sim \mathcal{N}(\phi_{prior}(o_t),\sigma^2 I)
% \end{equation}
We model the latent variable's prior distribution $p(z_t|o_t)$ and posterior 
 distribution $q(z_t|o_t,\textbf{\textit{q}}_t)$ as Gaussian distribution:
\begin{gather}
    p(z_t|o_t) \sim \mathcal{N}(\pi_{\text{prior}}(z_t|o_t), \sigma^2 I), \notag\\
    q(z_t|o_t,\textbf{\textit{q}}_t) \sim \mathcal{N}  (\pi_{{\text{MT}}}( z_t|o_t,\textbf{\textit{q}}_t)+\pi_{\text{prior}}(z_t|o_t), \sigma^2 I),
\end{gather}
where $\pi_{{\text{MT}}}(z_t|o_t,\textbf{\textit{q}}_t)$ and $\pi_{\text{prior}}(z_t|o_t)$ are neural networks parameterized by $\theta_{\text{MT}}$ and $\theta_{\text{prior}}$.  $I$ is a identity matrix, and $\sigma$ is a fixed standard deviation for simplicity. 
% Then the policy net can produce the action by:
% \begin{equation}
%     a_t=\phi_\pi(o_t,z_t)
% \end{equation} Consequently, we can reconstruct the next state $\hat{s}_{t+1}$ using Eq. \ref{eq:prediction}. 

% This task can be likened to an auto-encoder problem, where we encode the target trajectory $\hat{\tau}$ into the latent space $z$ and then reconstruct the real trajectory $\tau$ using the policy network and word model. 

The tracking learning loss is defined as follows:
\begin{equation}
     L^{\text{T}}_t = 0.6L^{\text{jpos}}_t+0.05L^{\text{jvel}}_t+0.3L^{\text{bpos}}_t+0.05L^{\text{bvel}}_t,
\end{equation}
% \begin{gather}
%     L_{\text{IL}} = \frac{1}{n}\sum_{t=1}^{n}\alpha^t{Dis(s_t,q_{t-1})}\notag \\
%     s.t. \quad s_{t+1} = \theta_{wm}(o_t,a_t)+s_t, a_t=\phi_\pi(o_t,z_t)
% \end{gather}
where the joint position loss $L^{\text{jpos}}_t$, joint velocity loss $L^{\text{jvel}}_t$, base position loss $L^{\text{bpos}}_t$ and base velocity loss $L^{\text{bvel}}_t$ are similar to the reward function in~\cite{li2023learning}:
\begin{gather}
    L^{\text{jpos}}_t = 1-\exp(-{\lVert \hat{j}_t-\bar{j}_t\rVert}^2 ), \notag \\
    L^{\text{jvel}}_t = 1-\exp(-{\lVert \hat{\dot{j}}_t-\bar{\dot{j}}_t\rVert}^2 ), \notag \\
    L^{\text{bpos}}_t = 1-\exp(-20{\lVert \hat{p}^{\text{base}}_t-\bar{p}^{\text{base}}_t\rVert}^2-10{\lVert \hat{i}^{\text{base}}_t-\bar{i}^{\text{base}}_t\rVert}^2 ), \notag \\
    L^{\text{bvel}}_t = 1-\exp(-2{\lVert \hat{\dot{p}}^{\text{base}}_t-\bar{\dot{p}}^{\text{base}}_t\rVert}^2-0.2{\lVert \hat{\dot{i}}^{\text{base}}_t-\bar{\dot{i}}^{\text{base}}_t\rVert}^2 ),
\end{gather}
where $j$ and $\dot{j}$ are joint position and joint velocity, $p^{\text{base}}$ and $i^{\text{base}}$ represent base position and orientation, $\dot{p}^{\text{base}}$ and $\dot{i}^{\text{base}}$ denote base velocity and base angular velocity. $\hat{(\cdot)}$ represents the states predicted by the world model and $\bar{(\cdot)}$ denotes the reference motion. 
Since the world model is differentiable, the gradient of the motion loss can be calculated end-to-end through the differential dynamics. 

To ensure the latent space is well formed so that we can further find an appropriate latent variable in the downstream command following task, we incorporate a KL-divergence loss for regularization:
\begin{equation}
\begin{split}
    L^{\text{KL}}_t &= {D_{\text{KL}}(q(z_t|o_t,\textbf{\textit{q}}_t)||p(z_t|o_t))} \\
                    &={\lVert \pi_{{\text{MT}}}(o_t,\textbf{\textit{q}}_t)\rVert}^2/2\sigma^2.
\end{split}
\end{equation}
To make our policy focus on not only the next state but also a long-term horizon, the final loss term is calculated as the sum of n-step tracking loss, where the n-step roll out is predicted by the world model
\begin{equation}
    L^{\text{MT}}_t = \sum_{t=1}^n(L^{\text{T}}_t+0.1L^{\text{KL}}_t).
\end{equation}

\subsection{Command Following}
The next step is to train a policy that follows linear velocity and angular velocity specified by users. We introduce the Command Following Encoder $\pi_{\text{CF}}(z_t|o_t, c_t)$ to encode the commands into the latent space. Given a random command $c_t=[\bar{v}_t,\bar{\omega}_t]$, where $\bar{v}_t$ and $\bar{\omega}_t$ represent the desired linear velocity in forward direction and the desired angular velocity, the posterior distribution of latent variable $z_t$ is computed as:
\begin{equation}
    q(z_t|o_t,c_t) \sim \mathcal{N}(\pi_\text{prior}(z_t|o_t)+\pi_{\text{CF}}(z_t|o_t,c_t),\sigma^2 I),
\end{equation}
where $\pi_\text{CF}(z_t|o_t,c_t)$ is the neural network parameterized by $\theta_\text{CF}$. Since our goal is to make the robot follow the command, the command following loss comprises both linear velocity loss $L^{\text{v}}_t$ and angular velocity loss $L^{\omega}_t$:
\begin{gather}
    L^{\text{CF}}_t = 2L^{\text{v}}_t+L^{\omega}_t,
    \label{eq:track_loss}\\
% \end{equation}
% \begin{equation}
    L^{\text{v}}_t = 1-\exp(-2|\bar{v}_t-\hat{v}_t|), \\
% \end{equation}
% \begin{equation}
    L^{\omega}_t = 1-\exp(-2|\bar{\omega}_t-\hat{\omega}_t|),
\end{gather}
where $\bar{(\cdot)}$ represents the user command while $\hat{(\cdot)}$ is the robot states predicted by the world model. 
To preserve the naturalness of the robot behavior, we exclusively update the command following network $\pi_{\textbf{CF}}$ while keeping the prior network $\pi_{\text{prior}}$ and motor decoder $\pi_M$ fixed during training. 

% We first train the tracking policy from scratch with random commands. 

% Given the robot's current position $p^{base}$ and a sequence of future path way points $P = [p_1, ..., p_n]$, we initially locate a target point from the path using a specific lookahead distance $l_{ahead}$:
% \begin{equation}
%     \hat{p}=\mathop{\arg\min}_{p_i,p_i\in P} \ \ | \| p^{base}-p_i\|-l_{ahead}|
% \end{equation}
% where the look ahead distance is correlation with the target speed that $l_{ahead}=a\hat{v}^x$, and $a$ is a hyperparameter.  With the target point determined, we can compute the target yaw angle as follows:
% \begin{equation}
%     \hat{\theta}=arctan(\frac{\hat{p}_y-p^{base}_y}{\hat{p}_x-p^{base}_x})
% \end{equation}
% Given the robot's current yaw angle $\theta^{base}$, the target yaw angular velocity of the command is based on the error in yaw angle:
% \begin{equation}
%     \hat{\omega} = k_p*(\hat{\theta}-\theta^{base})
% \end{equation}
% where $k_p$ is a hyperparameter for proportional control. \tgli{Whether or not discuss pure pursuit algorithm should be further discussed, since it's not a contribution of this work.}

\subsection{Fine-tune on a Real Robot}
Owing to the sim-to-real gap, the policy learned from the simulation may fail when deployed on the real robot. Hence, we fine-tune both the Command Following Encoder and the Motor Decoder on the real robot to follow the desired paths.
% Owing to the sim2real gap, the policy network trained in simulation may not exhibit proficient motor skills on a real robot. Consequently\shj{Since the motor decoder trained in the simulator may not exhibit proficient motor skills on a real robot}, we update both the command and policy networks during the fine-tuning phase. 
To preserve the natural behavior originating from the original Motor Encoder $\pi_{\text{M}}^{ori}$, we introduce a regularization term: 
\begin{equation}
    L^{\text{reg}}_t = \lVert \pi^{\text{ori}}_{\text{M}}(a_t|o_t,z_t)- \pi_{\text{M}}(a_t|o_t,z_t) \rVert.
    \label{eq:real_reg}
\end{equation}
The algorithm for fine-tuning in the real world is outlined in Algorithm~\ref{alg:fine-tune}.
\begin{algorithm}[t]
\caption{Fine-tune on the real robot}
\label{alg:fine-tune}
\begin{algorithmic}[1]
    \REQUIRE Network parameters $\theta_{\text{w}},\theta_{\text{prior}},\theta_{\text{CF}},\theta_{\text{M}}$, learning rates $\alpha_{\text{w}},\alpha_{\pi}$, update number $n_{\text{w}},n_{\pi}$, max iterations $N$, samples $n_{\text{sample}}$, rollout length $n$, replay buffer $\mathcal{B}$, batch size $M$
    \FOR{i in \{0...N-1\}}
    \STATE send updated control policy to the real robot
    \STATE roll out $n_{\text{sample}}$ steps on the real robot, send to $\mathcal{B}$
    \tcp{Update the world model}
    \FOR{j in \{1...$n_{\text{w}}$\}}
        \STATE sample trajectories ${(s_0,a_0,...,s_n,a_n)}_{1...M}$ from $\mathcal{B}$
        \STATE compute prediction loss $L^{\text{w}}$ in Eq.~\ref{eq:prediction_loss}
        \STATE $\theta_{\text{w}}\leftarrow \theta_{\text{w}}+\alpha_{\text{w}}\nabla_{\theta_{\text{w}}}{L^{\text{w}}}$
    \ENDFOR
    
    \tcp{Update the control policy}
    \FOR{j in \{1...$n_{\pi}$\}}
        \STATE sample states and commands ${(s_0,c_0)}_{1...M}$ from $\mathcal{B}$
        \STATE roll out $n$ steps predicted by world model $f_{{\text{w}}}$
        \STATE compute $L^{\textbf{CF}} \leftarrow \sum_{t=1}^n (L^{\text{CF}}_t+0.1L^{\text{reg}}_t$) in Eq. \ref{eq:track_loss}, \ref{eq:real_reg}
        \STATE $\theta_{\text{CF}}\leftarrow \theta_{\text{CF}}+\alpha_{\pi}\nabla_{\theta_{\text{CF}}}{L^{\textbf{CF}}}$
        \STATE $\theta_{\text{M}}\leftarrow \theta_{\text{M}}+\alpha_{\pi}\nabla_{\theta_{\text{M}}}{L^{\textbf{CF}}}$
    \ENDFOR    
    \ENDFOR
\end{algorithmic}
\end{algorithm}
% \vspace{-6em}
% \begin{figure}[h]
% 	% \vspace{-.3cm}
% 	\centering
% 	\includegraphics[width=0.5\textwidth]{Simu_efficient.png}
% 	\caption{The training curve of our method and PPO algorithm in the simulation on training tracking task from scratch. Each method conducts three experiments to derive the mean and standard deviation of the mean rewards.}
% 	\label{fig:track_loss}
% 	% \vspace{-.4cm}
% \end{figure}
% \begin{figure}[h]
% 	\vspace{-.3cm}
% 	\centering
% 	\includegraphics[width=0.5\textwidth]{Simu_adapt1.png}
% 	\caption{ Training curve of our method and PPO algorithm in simulation for fine-tuning policy in the changed environment.}
% 	\label{fig:simu_adapt1}
% 	\vspace{-.4cm}
% \end{figure}
% \begin{figure}[h]
% 	\vspace{-.3cm}
% 	\centering
% 	\includegraphics[width=0.5\textwidth]{Simu_adapt2.png}
% 	\caption{ Mean loss in three workloads 3kg, 5kg, and 7kg with control latency equal to 6ms in the simulation for adaptation on the path-following task, where loss equals to $2e_{speed}+e_{\Omega}$.}
% 	\label{fig:simu_adapt2}
% 	\vspace{-.4cm}
% \end{figure}
\begin{figure}[t]
\flushleft
	\vspace{-.1cm}
	% \centering
	\includegraphics[width=0.45\textwidth]{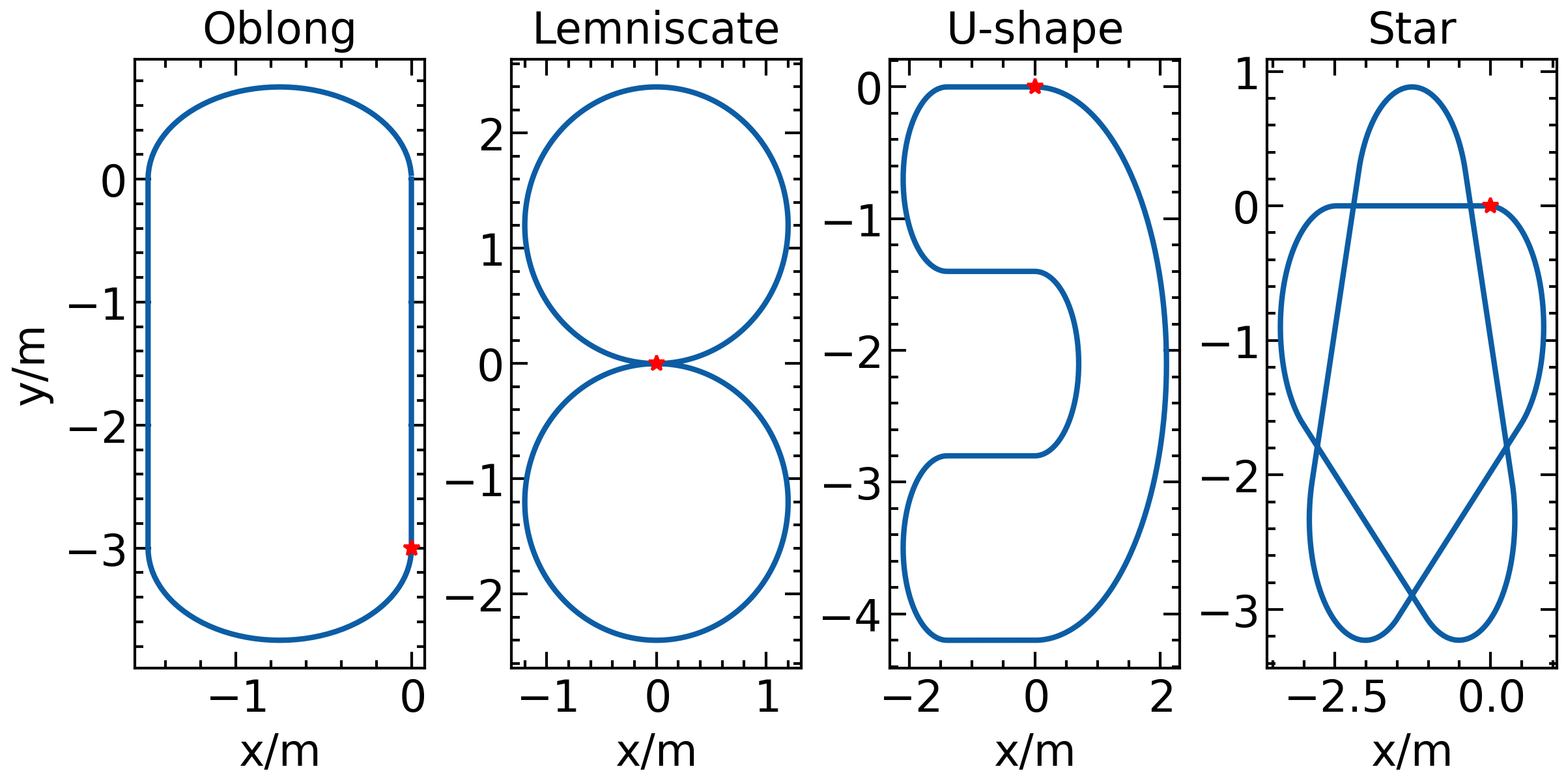}
	\caption{ Four types of desired paths. The red star represents the starting point.}
	\label{fig:paths}
	\vspace{-.4cm}
\end{figure}

\section{EXPERIMENT RESULTS}
In this section, we report experimental results to address the following pivotal questions: 
(i) How effective is our approach in improving sample efficiency, compared with RL methods?
(ii) How well is our fine-tuning process on the real robot can help to close the sim-to-real gap?
(iii) Does our fine-tuned policy exhibit sufficient generalization capacity on previously unseen tasks?
We conduct experiments both in simulation and real world. We compare our method with a RL baseline in terms of sample efficiency. \shj{For motion tracking, our problem setting is different from imitation learning since we only have the target motion states and lack the groundtruth actions executed on each joint motor. Therefore, we do not compare with imitation learning algorithms.} In the real world experiment, we conduct the fine-tuning process on a real quadrupedal robot. To further demonstrate the generalization ability, we performance path following task on four unseen paths.

% To assess the efficacy of our methods, we conducted experiments in two settings: the Isaac Gym simulation environment \cite{makoviychuk2021isaac} and on an actual quadrupedal robot. Our experiments are designed to address the following pivotal questions:
%     \begin{enumerate}
% 	\item Can our approach attain a high level of sample efficiency during both training and fine-tuning?
% 	\item Can our approach enhance the performance of the policy on the real robot through fine-tuning with a limited number of samples?
% 	\item Does our fine-tuned policy exhibit sufficient generality to perform effectively on previously unseen tasks?
% 	\end{enumerate}
 
\subsection{Evaluation in Simulation Environments}
\begin{figure}[t]
	% \vspace{-.3cm}
	\centering
	\includegraphics[width=0.45\textwidth]{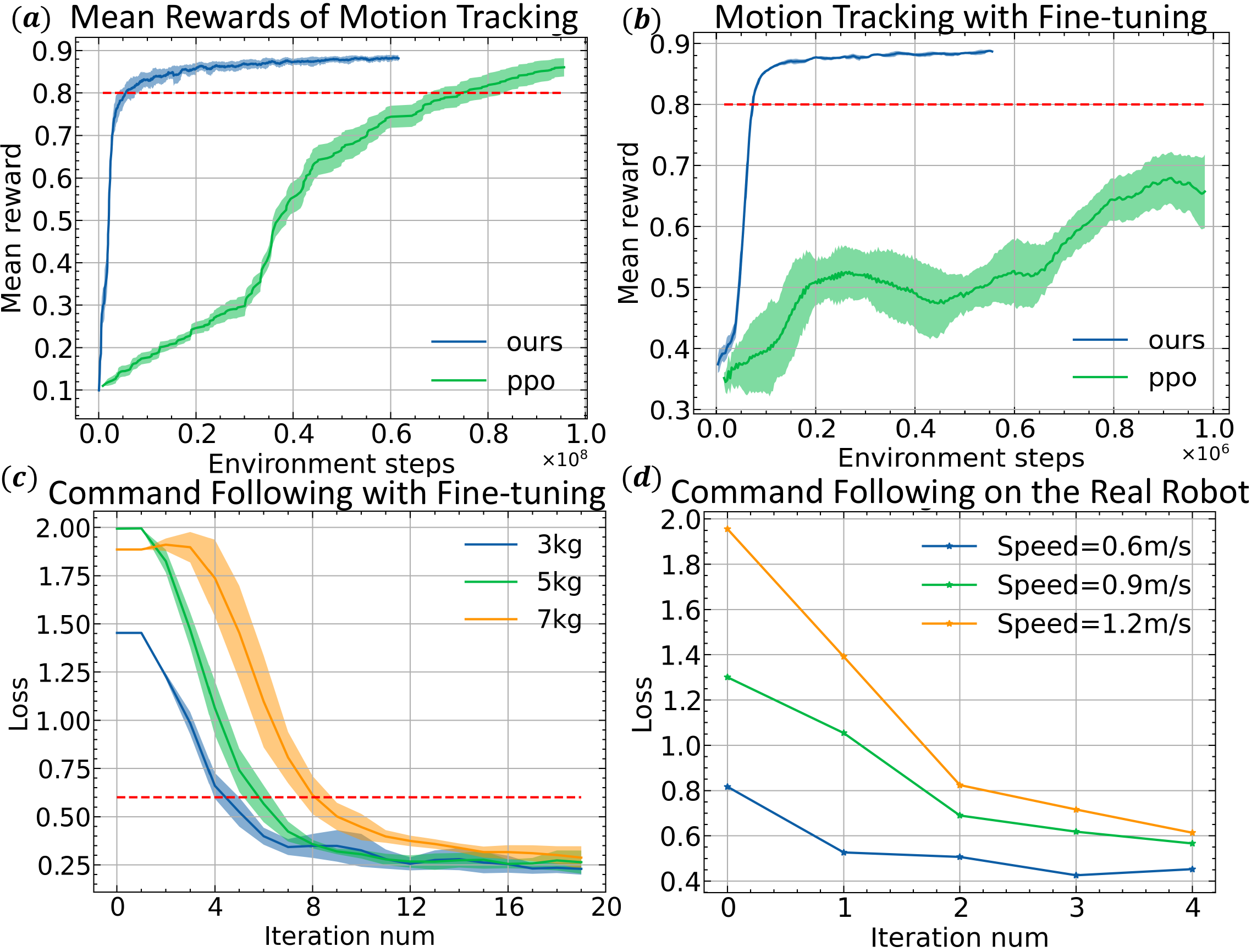}
	\caption{(a) Training curves of the motion tracking task in the simulation. (b) Training curves of fine-tuning the motion tracking task policy in the modified simulation environment. (c) Mean loss of fine-tuning the path following policy in three workloads within the modified simulation environment. (d) Mean loss of fine-tuning the path following policy under various speeds on the real robot.}
	\label{fig:train_curve}
	\vspace{-.2cm}
\end{figure}

\textbf{Sample Efficiency in Motion Tracking.}
To address the first question regarding sample efficiency, we initially train the motion tracking task from scratch using Isaac Gym. Isaac Gym is a GPU-based physical simulator simulating a batch of agents concurrently. In this task, we simultaneously employ 128 agents for training. We compare our method with PPO algorithm~\cite{schulman2017proximal} with respect to the number of samples collected from the simulator. The reward function for PPO is defined as $r_t=1-L_t^{\text{I}}$. We maintain an identical policy network structure for both methods to facilitate a meaningful comparison. The mean reward during training is reported in Fig.~\ref{fig:train_curve}(a). It demonstrates that our method achieves a mean reward of 0.8 with approximately 5 million samples, as indicated in the read dashed line. In contrast, the PPO algorithm requires over 70 million samples to achieve similar results. This showcases that our method's sample efficiency surpasses that of PPO by over tenfold.

\textbf{Sample Efficiency in Adapting to New Environments.}
Directly training PPO on a real robot is dangerous and may easily damage the robot. To compare the sample efficiency in adapting to new environments, we introduce variations to physical parameters in simulation and perform the fine-tuning process to make the adaptation. For the motion tracking task, we alter a number of physical parameters as shown in \textit{Env1}, TABLE~\ref{tab:physical_parameters}. For example, we significantly increase the robot's mass from 5.74 kg to 14 kg, which makes the new environment extremely difficult for the original policy. To emulate a scenario akin to real-world robot data collection, we employ 2 agents in the simulation environment for both methods. In our approach, each training iteration accumulates 3000 samples, equivalent to 1 minute of data collection given the control frequency of 50 Hz. For PPO, policy updates are conducted every 32 steps. Fig.~\ref{fig:train_curve}(b) depicts the training curves. The plot highlights that our method attains a mean reward of 0.8 with roughly 50,000 samples (equivalent to approximately 17 minutes of data) in this challenging setting. Conversely, PPO algorithm remains subpar even with ten times the sample size.
\begin{table}[]
\centering
\caption{The physical parameters of the original and new environments, where Ctrl Lat represents Control Latency.}
\label{tab:physical_parameters}
\begin{tabular}{c|cccc}
\hline
            & Mass (kg) & Kp    & Ctrl Lat (ms) & Max Torque (Nm)   \\ \hline
Original    & 5.74      & 50.0  & 0.0           & 18.0              \\
Env1        & 14.0      & 40.0  & 6.0           & 16.2              \\ 
Env2        & 5.74+3.0       & 50.0  & 6.0           & 18.0              \\ 
Env3        & 5.74+5.0       & 50.0  & 6.0           & 18.0              \\ 
Env4        & 5.74+7.0       & 50.0  & 6.0           & 18.0              \\ \hline
\end{tabular}
\vspace{-.2cm}
\end{table}

% Additionally, we conduct experiments on fine-tuning command following policy. 
% To further investigate the model performance, we extend the task of command following to path following. Given a sequence of path way points in Cartesian coordinate, the objective of path following is to find the corresponding sequence of control command such that minimize the distance between the desired path and the robot real trajectory. An accurate command following model would contribute to the path following performance. Here we employ the pure pursuit algorithm~\cite{coulter1992implementation} to generate real-time commands for path following. 
To further investigate the performance of command following, we extend this task to path following, where the robot aims at following predefined paths, as shown in Fig.~\ref{fig:paths}. We employ the pure pursuit algorithm~\cite{coulter1992implementation} to convert the path information to commands. In this experiment, we follow the \textit{Oblong} with a target speed of 0.9 m/s. We also create three distinct environments, \textit{Env2}, \textit{Env3}, \textit{Env4}, as shown in TABLE~\ref{tab:physical_parameters}. To emulate the fine-tuning process in the real-world environment, each training iteration involves collecting 1500 samples (30 seconds data). Fig.~\ref{fig:train_curve}(c) depicts the training curves with the loss term defined in Eq. \ref{eq:track_loss}. 
From the plot, we observe that our approach, under workloads of 3kg, 5kg, and 7kg,requires approximately 4 iterations (2 minutes), 6 iterations (3 minutes), and 8 iterations (4 minutes) of data to achieve a loss of less than 0.6. This result indicates a relatively good performance at these speeds. In comparison, the loss of PPO remains nearly unchanged with such a limited amount of samples, and thus we did not draw the result. In this way, we can demonstrate the high sample efficiency of our approach for both training and fine-tuning, adapting to different environments in both motion tracking task and path following tasks.

% \begin{figure}[h]
% 	\vspace{-.3cm}
% 	\centering
% 	\includegraphics[width=0.5\textwidth]{Realrobot_losses.png}
% 	\caption{ Training curve of fine-tuning for four iterations with target speeds equal to 0.6m/s, 0.9m/s, and 1.2m/s on real robots. The loss contains speed tracking loss and angular velocity tracking loss.}
% 	\label{fig:real_loss}
% 	\vspace{-.3cm}
% \end{figure}

% \begin{figure}[h]
% 	% \vspace{-.3cm}
% 	\centering
% 	\includegraphics[width=0.48\textwidth]{speedfollow.pdf}
% 	\caption{(a) Speed following at 1.2 m/s along the oblong path on the real robot with real-world adaptation. (b) Speed following along the oblong path on the real robot using the original policy and the adapted policy.}
% 	\label{fig:real_speed12}
% 	\vspace{-.4cm}
% \end{figure}
\begin{figure}[t]
	% \vspace{-.3cm}
	\centering
	\includegraphics[width=0.4\textwidth]{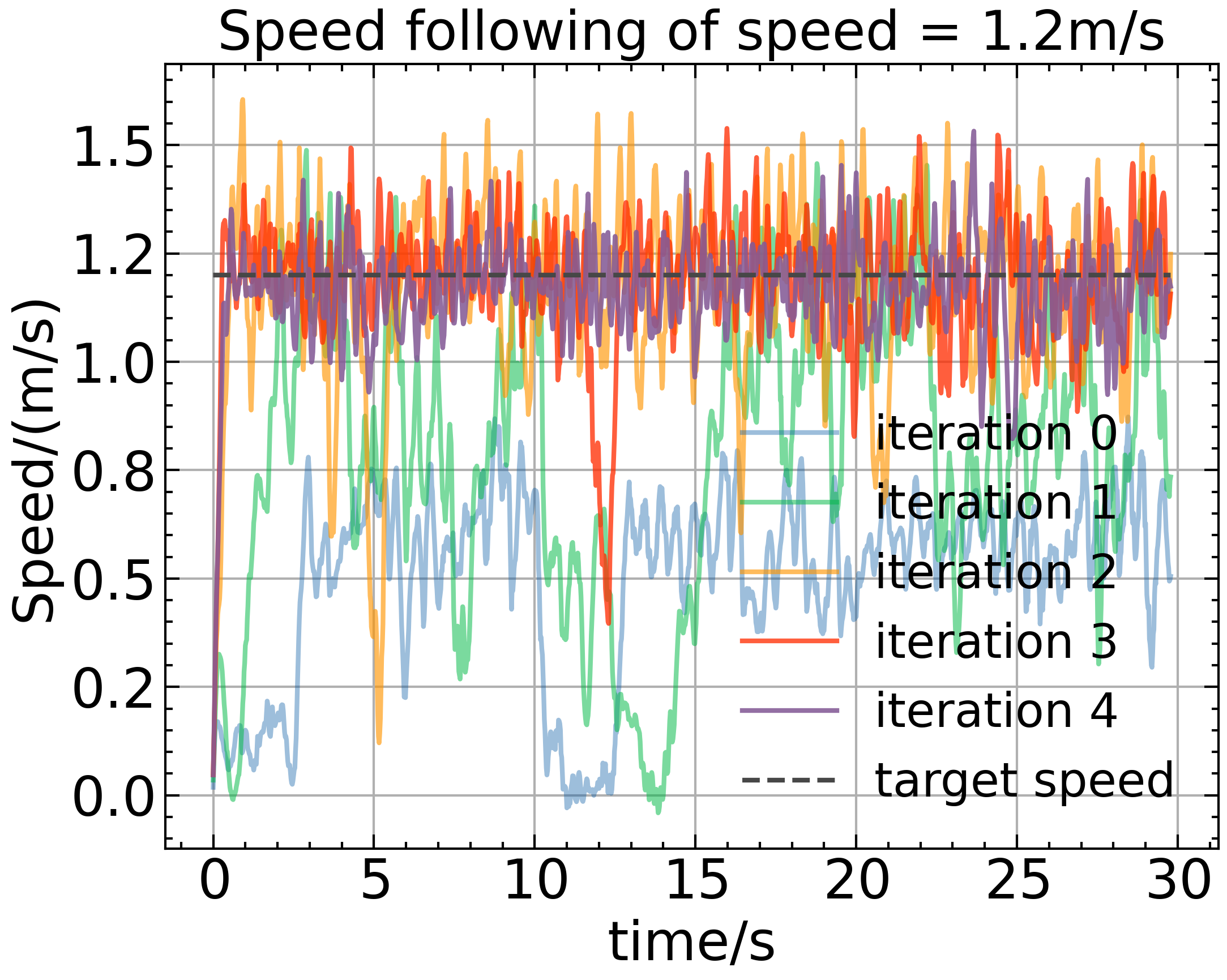}
	\caption{ Speed following at 1.2 m/s along the oblong path on the real robot with real-world adaptation.}
	\label{fig:real_speed12}
	\vspace{-.2cm}
\end{figure}
\begin{figure}[t]
	% \vspace{-.3cm}
	\centering
	\includegraphics[width=0.4\textwidth]{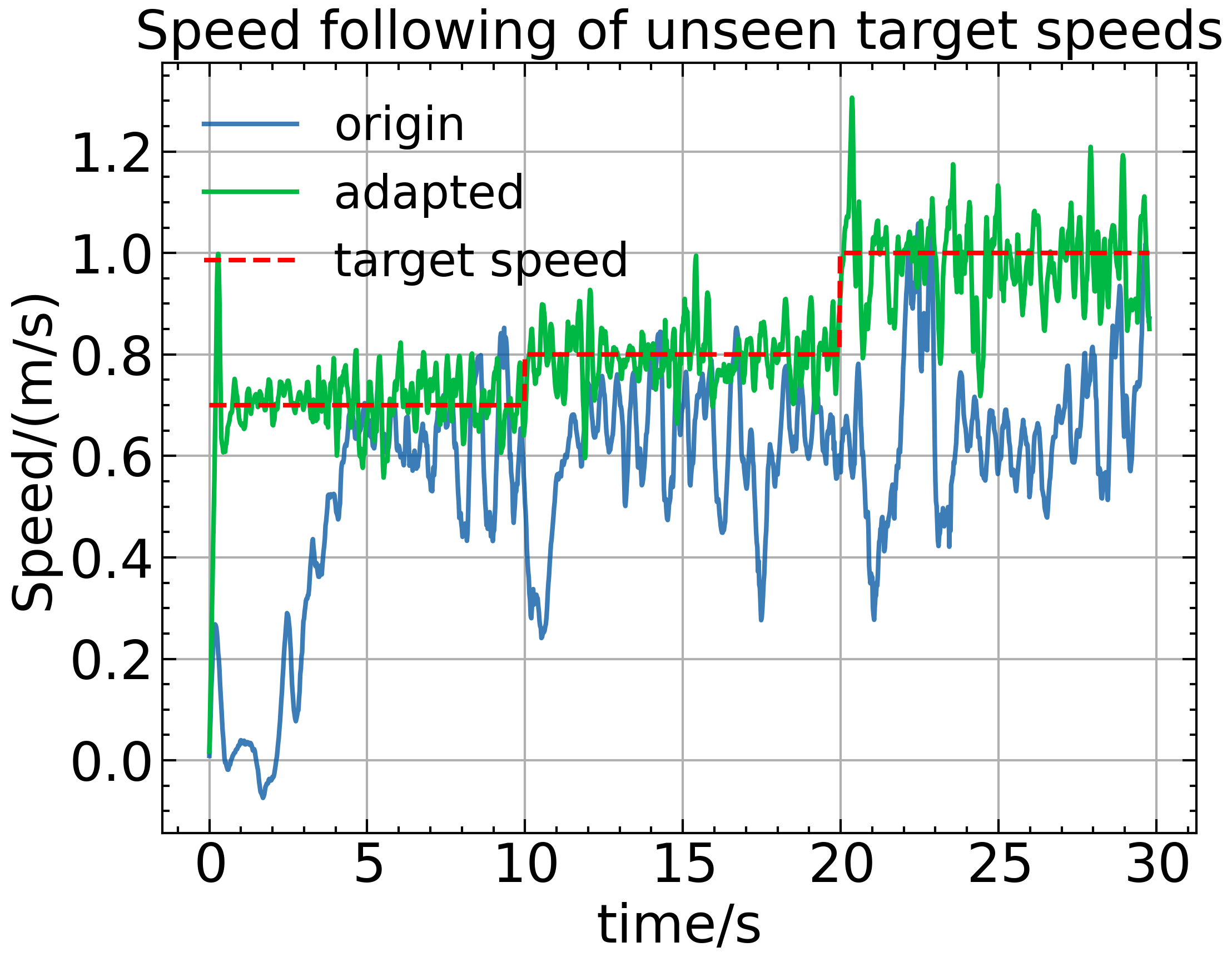}
	\caption{ Speed following along the oblong path on the real robot using the original policy and the adapted policy.}
	\label{fig:real_speedstar}
	\vspace{-.4cm}
\end{figure}
\subsection{Evaluation on Real World Experiments}
\textbf{Adapting from Simulation to Reality.}
To address the second question, we perform physical experiments using the real robot Max. 
% \tgli{Add description why we need fine-tune on real robot, instead of directly deploy the policy. Mention sim2real gap here.}
Due to the sim2real gap, the policy trained in the simulation may fail to follow the path with the desired speed and can exhibit significant lag behind the target speed, especially at high target speeds. This underscores the necessity of real-world fine-tuning. 
% Our objective is to fine-tune the policy on the real robot, enabling it to follow commands generated by the high-level controller and thereby track the path on the playground as illustrated in Fig.~\ref{fig:paths}(d). 
We perform three adaptation experiments on \textit{Oblong} with target speeds of 0.6m/s, 0.9m/s, and 1.2m/s. To fine-tune the policy in the real world, each iteration involves collecting 30 seconds of data (1500 samples) to train the world model, followed by updating the policy network using data predicted by the adapted world model. Fig.~\ref{fig:train_curve}(d) displays the command following loss $L^{\text{CF}}$ for four iterations (2 minutes data) with target speeds of 0.6m/s, 0.9m/s, and 1.2m/s on the real robot. TABLE~\ref{tab:real_adaptation} presents the averaged linear velocity error $e_v=\frac{1}{n}\sum_{t=1}^n|\bar{v}_t-\hat{v}_t|$ and angular velocity loss $e_{\omega}=\frac{1}{n}\sum_{t=1}^n|\bar{\omega}_t-\hat{\omega}_t|$ computed within a 30-second trajectory after each iteration during the real-world adaptation. It is evident that after the first iteration, there is a significant decreasing in losses. Particularly for a speed of 1.2m/s, the speed error decreases by more than 0.26m/s. After four iterations, the losses appear to converge, and the final performance is highly effective in tracking the commands. For example, Fig.~\ref{fig:real_speed12}(a) depicts speed tracking at 1.2m/s on the real robot with real-world adaptation. It is evident that in the initial policy (iteration 0), the actual speed lags considerably behind the target speed. After the first iteration, the actual speed can somewhat follow the target, but it exhibits significant fluctuations. In iteration 4, the policy effectively tracks the target speed with minimal vibration.
\begin{table}[]
% \vspace{-.3cm}
\centering
\caption{The averaged linear velocity error $e_v$ and angular velocity loss $e_{\omega}$ computed in a trajectory of 30s after each iteration. Iter0 refers to the original policy without any fine-tuning.}
\label{tab:real_adaptation}
\begin{tabular}{c|cccccc}
\hline
      & \multicolumn{2}{c}{Speed=0.6m/s} & \multicolumn{2}{c}{Speed=0.9m/s} & \multicolumn{2}{c}{Speed=1.2m/s} \\ \hline
      & $e_v$           & $e_{\omega}$          & $e_v$           & $e_{\omega}$          & $e_v$           & $e_{\omega}$          \\ \hline
Iter0 & 0.088           & 0.587          & 0.250           & 0.612          & 0.696           & 0.501          \\
Iter1 & 0.055          & 0.241          & 0.194           & 0.565          & 0.431          & 0.319          \\
Iter2 & 0.047           & 0.232          & 0.098           & 0.297          & 0.148           & 0.276          \\
Iter3 & \textbf{0.038}  & 0.190          & 0.078           & 0.269          & 0.103          & 0.286          \\
Iter4 & 0.047           & \textbf{0.189} & \textbf{0.063}  & \textbf{0.249} & \textbf{0.081} & \textbf{0.240} \\
\hline
\end{tabular}
\vspace{-.2cm}
\end{table}
\begin{table}[]
\centering
\caption{The averaged linear velocity error $e_v$, angular velocity error $e_{\omega}$, and distance error $e_{\text{p}}$ computed across four paths with unseen target speeds equal to 0.7m/s, 0.8m/s and 1.0m/s. }
\label{tab:real_general}
\begin{tabular}{ccccccc}
\hline
\multicolumn{1}{c|}{}        & \multicolumn{3}{c|}{Oblong} & \multicolumn{3}{c}{Lemniscate} \\ \hline
\multicolumn{1}{c|}{}        & $e_{v}$     & $e_{\omega}$    & $e_{\text{p}}$      & $e_{v}$   & $e_{\omega}$  & $e_{\text{p}}$    \\ \hline
\multicolumn{1}{c|}{origin}  & 0.269     & 0.578    & 2.031    & 0.224   & 0.642  & 2.190  \\
\multicolumn{1}{c|}{adapted} & 0.052     & 0.239    & 0.901    & 0.057   & 0.199  & 0.725  \\ \hline
                             &           &          &          &         &        &        \\ \hline
\multicolumn{1}{c|}{}        & \multicolumn{3}{c|}{U-shape}          & \multicolumn{3}{c}{Star}  \\ \hline
\multicolumn{1}{c|}{}        & $e_{v}$     & $e_{\omega}$    & $e_{\text{p}}$      & $e_{v}$   & $e_{\omega}$  & $e_{\text{p}}$    \\ \hline
\multicolumn{1}{c|}{origin}  & 0.233     & 0.572    & 1.560    & 0.287   & 0.631  & 2.031  \\
\multicolumn{1}{c|}{adapted} & 0.053     & 0.210    & 0.771    & 0.050   & 0.234  & 0.901  \\ \hline
\end{tabular}
\vspace{-.5cm}
\end{table}
% \begin{table*}[]
% \centering
% \caption{The mean of speed error $e_{speed}$, angular velocity error $e_{\Omega}$ and distance error $e_{dis}$ computed in the four paths under unseen target speeds equal to 0.7m/s, 0.8m/s and 1.0m/s. }
% \label{tab:real_general}
% \begin{tabular}{c|cccccccccccc}
% \hline
%         & \multicolumn{3}{c|}{Playground} & \multicolumn{3}{c|}{Eight} & \multicolumn{3}{c|}{U} & \multicolumn{3}{c}{Star} \\ \hline
%         & $e_{speed}$     & $e_{\Omega}$    & $e_{dis}$      & $e_{speed}$   & $e_{\Omega}$   & $e_{dis}$    & $e_{speed}$  & $e_{\Omega}$ & $e_{dis}$   & $e_{speed}$  & $e_{\Omega}$  & $e_{dis}$    \\ \hline
% origin  & 0.375     & 0.507    & 2.031    & 0.365   & 0.559   & 2.190  & 0.336  & 0.526 & 1.560 & 0.375  & 0.550  & 2.031  \\
% adapted & 0.094     & 0.326    & 0.901    & 0.090   & 0.311   & 0.725  & 0.095  & 0.304 & 0.771 & 0.094  & 0.289  & 0.901 
% \end{tabular}
% \end{table*}

\textbf{Generalization Ability on Unseen Scenarios.}
\begin{figure}[t]
	% \vspace{-.3cm}
	\centering
	\includegraphics[width=0.4\textwidth]{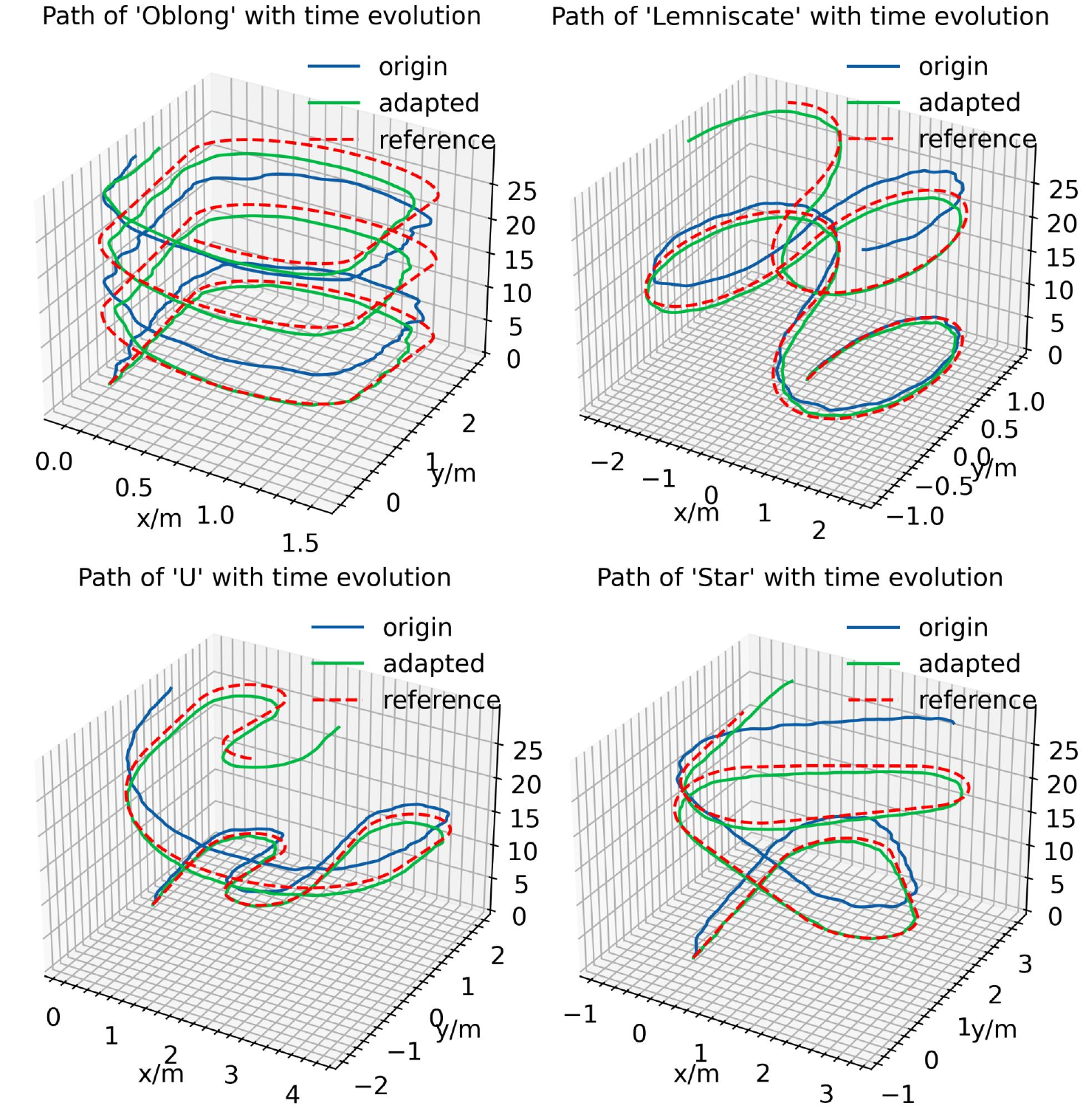}
	\caption{ Path following on unseen paths under the original policy and the adapted one. The z-axis represents the time evolution, and the reference path is computed by integrating the target speed with respect to time.}
	\label{fig:real_trajall}
	\vspace{-.5cm}
\end{figure}
To answer the last question, we evaluate our policy on unseen command velocities and paths. In the previous experiment, we collect real robot data, totaling 7.5 minutes of data with target speeds of 0.6m/s, 0.9m/s, and 1.2m/s. We utilize this data for off-policy fine-tuning to derive the adapted policy. 
We test the generalization ability on unseen target velocities of 0.7m/s, 0.8m/s, and 1.0m/s on all paths including unseen \textit{Lemniscate}, \textit{U-shape}, and \textit{Star}.
% with each target speed lasting 10s. Apart from the oblong path, we also conduct tests on three unseen paths depicted in Fig.~\ref{fig:paths}. 
TABLE~\ref{tab:real_general} reports averaged linear velocity error ($e_v$), angular velocity error ($e_{\omega}$), and distance error ($e_p$) computed over four paths lasting 30 seconds each. The distance error is defined as $e_p = \frac{1}{n}\sum_{t=1}^n\lVert p_t-p^*_t\rVert$,
% \begin{equation}
%     e_{dis} = \lVert p_t-p^*_t\rVert
% \end{equation}
where $p_t$, $p^*_t$ are robot position and the target position at time $t$. $p^*_t$ is derived by integrating the target speed with respect to time. 
From the table, it's evident that after off-policy adaptation, all of the errors have decreased by over one-half. 
% Particularly noteworthy is the speed tracking error, which is less than 0.1 for all scenes.
Fig.~\ref{fig:real_speed12} vividly demonstrates the speed tracking to follow the oblong path on the real robot under the original policy and the adapted one. The original policy lags behind the target unseen speeds, whereas our adapted policy can follow them effectively with averaged linear velocity error of around 0.05m/s. Fig.~\ref{fig:real_trajall} displays the real trajectories for tracking the path at different unseen target speeds. 
% The desired positions over time are computed by integrating the desired speed with respect to time. 
From the plot, it's evident that the original policy lags significantly behind the target trajectory, while our adapted policy can effectively track it, performing even slightly faster at higher speeds. In conclusion, the experimental results demonstrate that our adapted policy can successfully handle unseen commands and track unfamiliar paths, highlighting the generalization capability of our approach.

\section{CONCLUSIONS}

In summary, we have introduced an efficient learning framework designed to mimic the natural behavior of animals and enable path tracking for quadrupedal robots. Our approach begins by training a world model and a policy network, effectively turning it into an auto-encoder that utilizes the differential dynamics from the world model. This strategy significantly boosts sample efficiency, outperforming model-free deep reinforcement learning algorithms by over tenfold. Additionally, our method facilitates rapid policy fine-tuning on real robots, requiring only 2 minutes of data, and demonstrates robust generalization capabilities. Future directions could include developing a world model with perception information, allowing the framework to adapt to visual locomotion across challenging terrains. The fine-tuning algorithm can narrow the sim2real gap further and improve the success rate of visual locomotion in challenging environments. In conclusion, our work opens up exciting possibilities for training complex motor skills on real robots.

\addtolength{\textheight}{-2cm}   % This command serves to balance the column lengths

%%%%%%%%%%%%%%%%%%%%%%%%%%%%%%%%%%%%%%%%%%%%%%%%%%%%%%%%%%%%%%%%%%%%%%%%%%%%%%%%

%%%%%%%%%%%%%%%%%%%%%%%%%%%%%%%%%%%%%%%%%%%%%%%%%%%%%%%%%%%%%%%%%%%%%%%%%%%%%%%%

%%%%%%%%%%%%%%%%%%%%%%%%%%%%%%%%%%%%%%%%%%%%%%%%%%%%%%%%%%%%%%%%%%%%%%%%%%%%%%%

%%%%%%%%%%%%%%%%%%%%%%%%%%%%%%%%%%%%%%%%%%%%%%%%%%%%%%%%%%%%%%%%%%%%%%%%%%%%%%%%
{\small
	\bibliographystyle{ieeetr}
        % \balance
	\bibliography{myref}
}
\vfill

\end{document}